\documentclass[11pt]{article}

\usepackage[margin=1in]{geometry}
\usepackage{graphicx}
\usepackage{amsmath,amssymb}
\usepackage{siunitx}
\usepackage{booktabs,multirow}
\usepackage{authblk}
\usepackage{xcolor}
\usepackage[colorlinks=true,linkcolor=blue,citecolor=blue,urlcolor=blue]{hyperref}
\usepackage{float}
\usepackage[labelfont=bf,font=small]{caption}
\usepackage{subcaption}
\usepackage{microtype}          
\usepackage[section]{placeins}  
\graphicspath{{./}}             

\usepackage[nameinlink,capitalize]{cleveref}

\title{Solar Irradiation Forecasting using Genetic Algorithms}

\author[1]{V.~Gunasekaran\thanks{Correspondence: \href{mailto:vinodh.gunasekaran@circana.com}{vinodh.gunasekaran@circana.com}}}
\author[2]{KK.~Kovi}
\author[3]{S.~Arja}
\author[4]{R.~Chimata\thanks{Correspondence: \href{mailto:raghuveer.chimata@gattyinstruments.com}{raghuveer.chimata@gattyinstruments.com}}}

\affil[1]{Circana 203 N La Salle Dr STE 1500, Chicago, Illinois 60601, United States}
\affil[2]{AKHAN Semiconductor Inc, 940 Lakeside drive, Gurnee, Illinois 60031, United States}
\affil[3]{VzRAM Tech LLC, Lisle, Illinois, United States}
\affil[4]{GattyInstruments AB, Ulls V\"ag 29A, 75651 Uppsala, Sweden}

\date{}

\begin{document}
\maketitle

\begin{abstract}\noindent
Renewable energy forecasting is becoming increasingly important as its contribution to electrical power grids continues to grow. Solar energy, one of the most significant renewable resources, relies on accurate measurements of solar irradiation. To ensure effective grid management, high-precision forecasting models for solar irradiation are essential. In this study, machine learning techniques including Linear Regression, Extreme Gradient Boosting, and Genetic Algorithm Optimization are employed to forecast solar irradiation. Data for training and validation were collected from three geographically diverse stations in the United States that are part of the SURFRAD network. The models predict the Global Horizontal Index (GHI), which is used for performance comparison. Additionally, Genetic Algorithm Optimization is applied to the Extreme Gradient Boosting model to further enhance the accuracy of solar irradiation predictions.
\end{abstract}



\subsection*{Variables}
\begin{description}
  \item[dt] Decimal time.
  \item[zen] Solar zenith angle (\si{\degree}).
  \item[dw\_solar] Downwelling global solar.
  \item[uw\_solar] Upwelling global solar.
  \item[direct\_n] Direct-normal solar.
  \item[diffuse] Downwelling diffuse solar.
  \item[dw\_ir] Downwelling thermal infrared.
  \item[dw\_casetemp] Downwelling IR case temperature (K).
  \item[dw\_dometemp] Downwelling IR dome temperature (K).
  \item[uw\_ir] Upwelling thermal infrared.
  \item[uw\_casetemp] Upwelling IR case temperature (K).
  \item[uw\_dometemp] Upwelling IR dome temperature (K).
  \item[uvb] Global UVB.
  \item[par] Photosynthetically active radiation.
  \item[netsolar] Net solar (\(dw\_solar - uw\_solar\)).
  \item[netir] Net infrared (\(dw\_ir - uw\_ir\)).
  \item[totalnet] Net radiation (\(netsolar + netir\)).
  \item[temp] 10-meter air temperature (\si{\celsius}).
  \item[rh] Relative humidity (\si{\percent}).
  \item[windspd] Wind speed.
  \item[winddir] Wind direction (\si{\degree}, clockwise from north).
  \item[pressure] Station pressure (mb).
\end{description}

\section{Introduction}
The use of renewable energy worldwide has increased significantly in recent years. Although various renewable energy sources such as wind, tidal, and marine are available, solar energy has the potential to become the most significant renewable resource. Recent advances in technology have improved the efficiency of solar photovoltaic cells while reducing manufacturing costs~\cite{Khatib2012,Lazo2011,Dagne2013}. Accurate forecasting of solar irradiation is critical not only to optimize solar power generation but also to ensure effective grid management and identify alternative power sources when solar energy is unavailable. This is because solar irradiation is crucial for solar power generation. 

In recent studies, several machine learning (ML) models have been applied to predict solar irradiation, including Artificial Neural Networks~\cite{Yadav2014,Chow2012}, Probabilistic Models~\cite{Fatemi2018}, Bayesian Methods~\cite{Lauret2013}, Deep Learning Models~\cite{Alzahrani2017,Alzahrani2017b}, and Support Vector Machines~\cite{Wang2015}. In this work, we investigate and validate ML algorithms such as linear regression (LR) and extremal gradient boost (XGB)~\cite{Fan2018,Chen2016,Zhang2018}, along with genetic algorithm optimization (GA)~\cite{Holland1975,Goldberg1989,Skiena1998,Howard1995}. The data used in this study were obtained from three meteorological stations, Bondville, IL; Desert Rock, NV; and Penn State, PA, that are part of the SURFRAD network~\cite{Augustine2000,Wang2012}. These stations were selected for their diverse climatic conditions throughout the year, providing a comprehensive evaluation environment for the models.

The study focuses primarily on predicting the Global Horizontal Index (GHI), a key measure of solar irradiation, and evaluates the potential of Genetic Algorithms to enhance the forecasting accuracy of global solar irradiation. By integrating GA with traditional ML approaches such as LR and XGB, we aim to overcome the challenges associated with manual hyperparameter tuning and improve model robustness. The automated optimization process enabled by GA is particularly valuable in adapting to the nonlinear and dynamic nature of solar irradiation data, ultimately contributing to more reliable predictions.

This research not only demonstrates the superior performance of GA-optimized models but also highlights the broader implications of integrating advanced machine learning techniques into renewable energy forecasting. The findings suggest that the use of a GA approach can lead to significant improvements in prediction accuracy and computational efficiency. This, in turn, has the potential to facilitate better energy management and grid reliability, supporting the ongoing transition toward a more sustainable and resilient energy infrastructure.

\section{Data Preprocessing}
Solar irradiation data from the SURFRAD network, measured using a pyranometer, is available for the past 20 years from seven stations in different states of the United States. For this study, data from three stations, Bondville, IL; Penn State, PA; and Desert Rock, NV were selected due to their distinct climatic conditions throughout the year. The dataset covers three consecutive years from 2018 to 2020, with data from 2018--2019 used for training and data from 2020 used for validation and testing. This selection highlights the geographical variability in solar radiation. Only daytime data, recorded between 7 AM and 4 PM when solar irradiance is significant, were used, resulting in nine hours of data per day for model development.

\begin{figure}[!htbp]
    \centering
    \begin{subfigure}{0.97\linewidth}
        \centering
        \includegraphics[width=\linewidth]{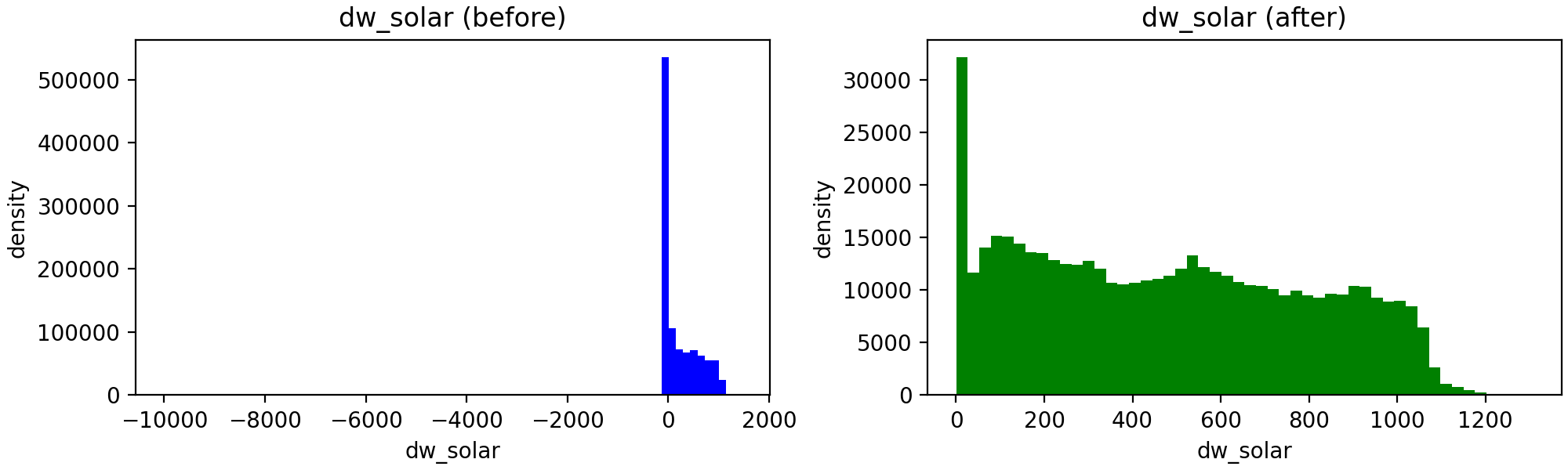}
        \caption{dw\_solar (before vs.\ after)}
    \end{subfigure}\vspace{0.5em}

    \begin{subfigure}{0.97\linewidth}
        \centering
        \includegraphics[width=\linewidth]{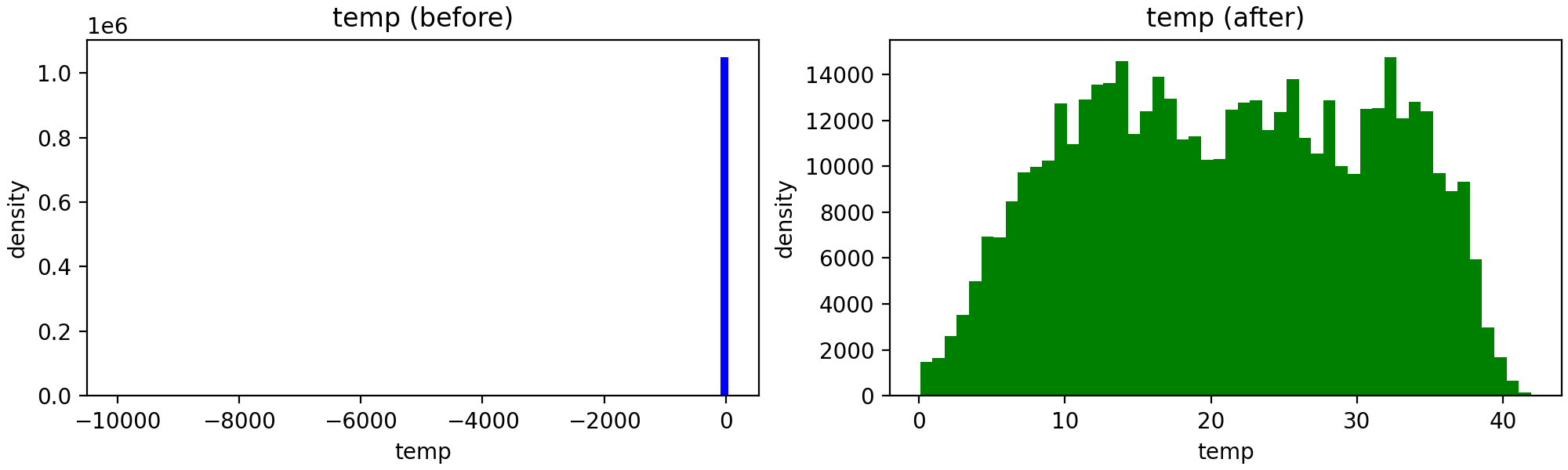}
        \caption{temp (before vs.\ after)}
    \end{subfigure}\vspace{0.5em}

    \begin{subfigure}{0.97\linewidth}
        \centering
        \includegraphics[width=\linewidth]{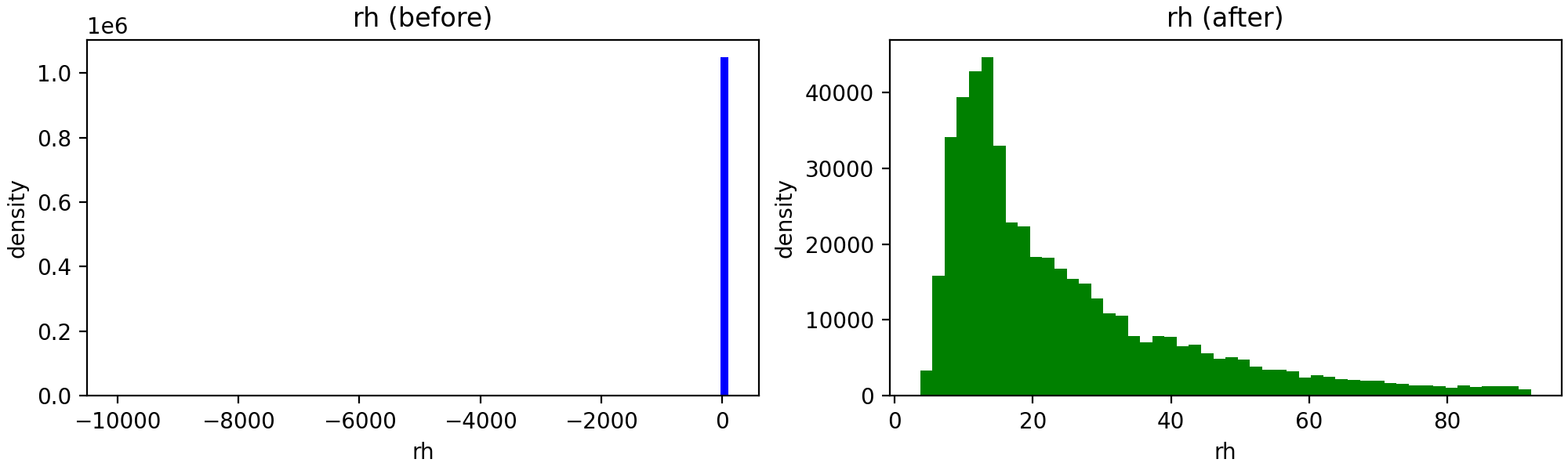}
        \caption{rh (before vs.\ after)}
    \end{subfigure}
    \caption{Data preprocessing to obtain normalized distributions by removing outliers and cleaning the data. Each row shows the variable before (left) and after (right) preprocessing.}
    \label{fig:preprocess}
\end{figure}

The models are designed to predict the Global Horizontal Index (GHI) for the next minute using input parameters such as temperature, pressure, wind speed, wind direction, relative humidity, solar zenith angle, net solar radiation, and time (detailed in minutes, hours, and months). Before training, the data were normalized, outliers were removed, and the data set was cleaned to ensure a normalized distribution. Few of the parameters can be seen in figure~\ref{fig:preprocess}, with plots showing an example of pre and post processed data.

\subsection{Outlier Detection}
Outliers were removed as they severely impact the functionality of the model. For example, values such as $-9999.90$ are recorded for at least 11 variables in the data. If replaced by mean, median, mode, or the minimum value, these values impact the results significantly. None of the four mentioned replacements work and hence were removed from the data. The fundamental statistics obtained from the data can be seen in Table~\ref{tab:summary-bondville}.

\section{Feature Selection}
\begin{figure}[!htbp]
    \centering
    \includegraphics[width=0.8\linewidth]{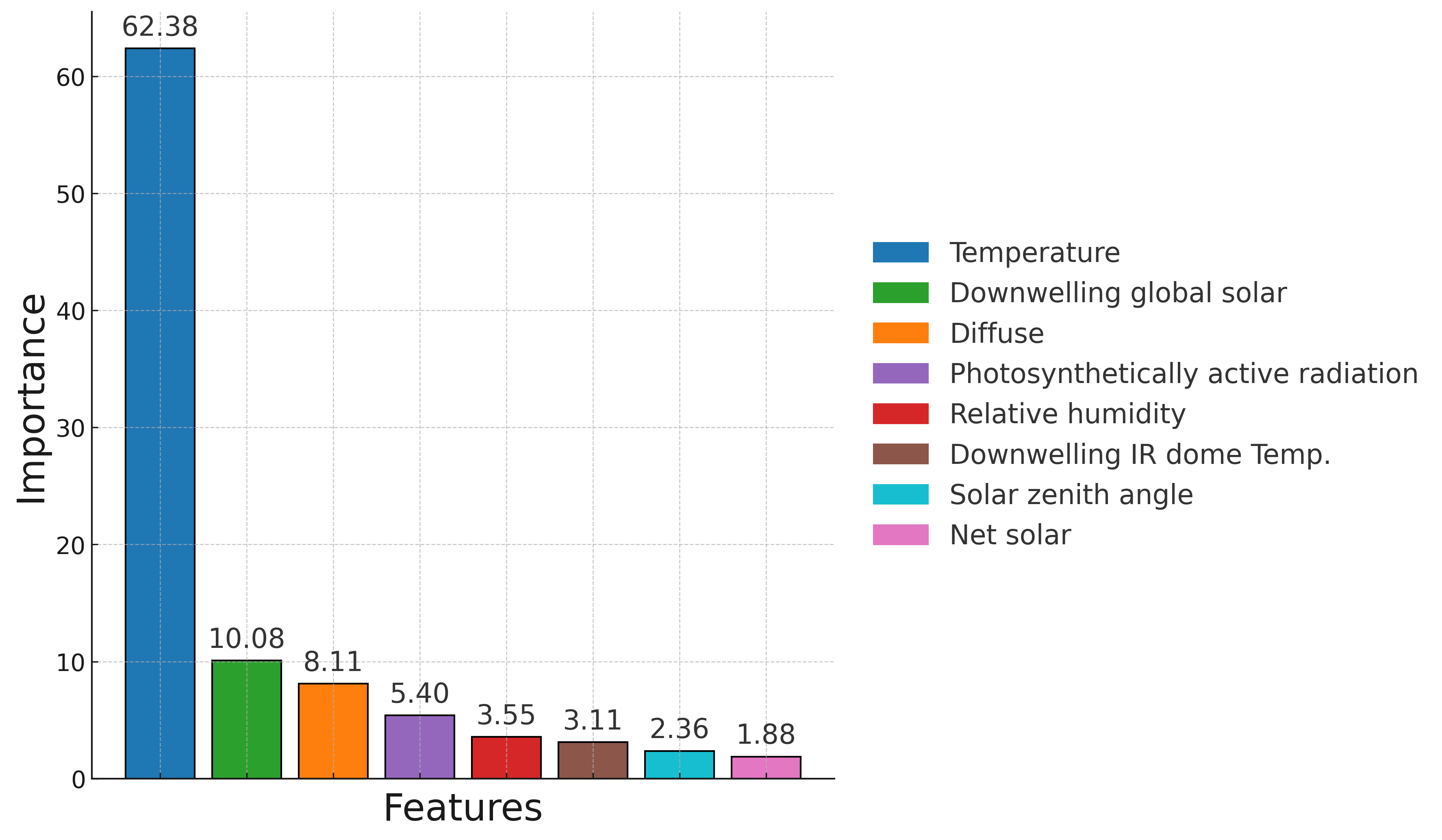}
    \caption{Feature importance used in the feature selection process (Random Forest). The top eight features are considered for the model.}
    \label{fig:featimp}
\end{figure}

Feature selection removes irrelevant features to enhance model performance by reducing both complexity and computational time. It also eliminates highly collinear variables. In this study, feature selection was performed by evaluating the importance of parameters using the Random Forest method. Out of fifteen variables, eight parameters that showed high relevance to the dependent variable were selected for training the model. Figure~\ref{fig:featimp} illustrates the features deemed most important.

\section{Machine Learning Algorithms}
\subsection{Linear Regression (LR)}
LR is one of the methods used in this study for predicting solar irradiation, where the dependent variable is continuous. LR models the relationship between the dependent variable and one or more independent variables by fitting a linear equation to the observed data. A simple LR equation is expressed as
\begin{equation}
    y = m x + b,
\end{equation}
where $y$ represents the predicted value, $x$ is the input variable, $m$ is the slope, and $b$ is the intercept.

\subsection{Extreme Gradient Boosting (XGB)}
XGB is a machine learning technique used for both regression and classification tasks. It constructs a predictive model by creating an ensemble of weak learners, typically decision trees, in a stage-wise manner. Like other boosting methods, XGB optimizes an arbitrary differentiable loss function~\cite{Chen2016}. It uses a partitioning algorithm to identify the optimal data split for a single target variable, and by resampling the data multiple times, it generates a weighted average from these resamples to form the final prediction. This approach, known as tree boosting, builds a series of decision trees into one robust predictive model. Similar to standalone decision trees, boosting does not assume any specific distribution for the data, yet it is less prone to overfitting because it gradually refines the model by combining multiple trees.

\subsection{Genetic Algorithm (GA)}
GA, first introduced by John Holland, is a meta-heuristic search and optimization algorithm inspired by Charles Darwin’s theory of natural selection. In GA, the best solutions are selected from a population and are combined and mutated to produce offspring that are progressively better. In this study, GA is employed to optimize the hyperparameters of the XGB model. This automated approach addresses the challenge of manual hyperparameter tuning, which can be laborious and may not always yield the best configuration for future predictions. By eliminating the need for blind selection, GA naturally improves the model’s accuracy without overfitting~\cite{Holland1975,Goldberg1989,Skiena1998,Howard1995}.

\subsection{Evaluation Metric}
The performance of these models is evaluated using the mean squared error (MSE), defined as
\begin{equation}
    \mathrm{MSE} = \frac{1}{N}\sum_{i=1}^{N}\left( y_i - (m x_i + b) \right)^2,
\end{equation}
where $N$ is the total number of observations (data points), $y_i$ is the actual value of an observation, and $(m x_i + b)$ is the prediction.

\begin{figure}[!htbp]
    \centering
    \includegraphics[width=0.8\linewidth]{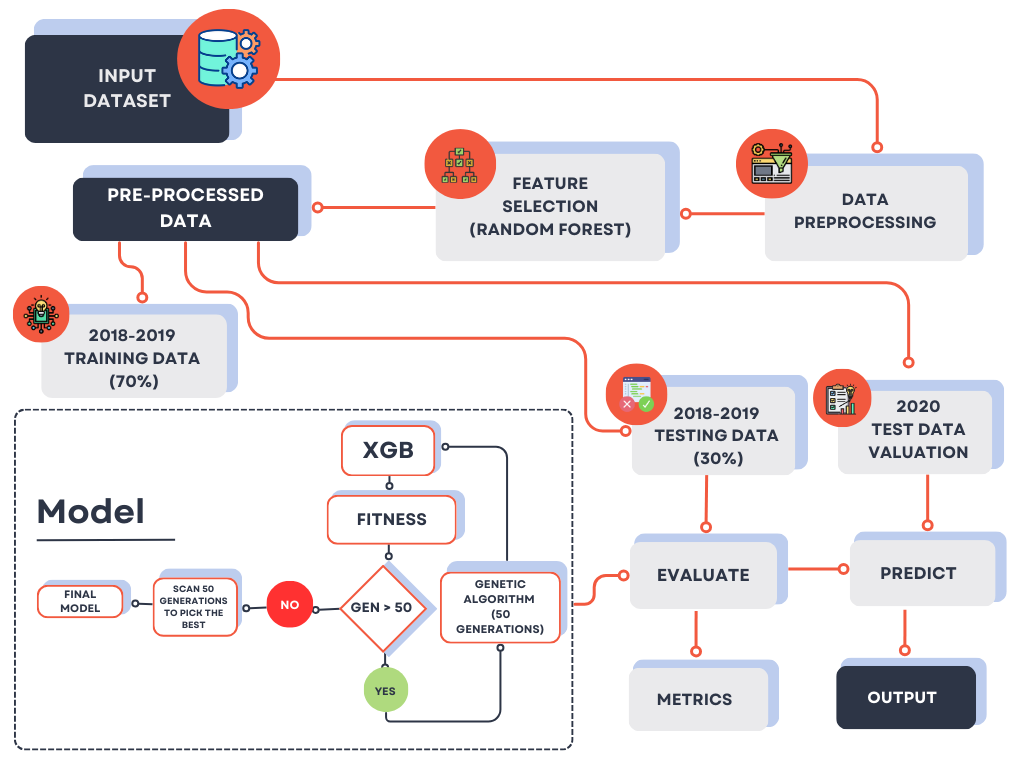}
    \caption{Flow chart of the training and validation process.}
    \label{fig:flowchart}
\end{figure}

\begin{table}[!htbp]
    \centering
    \caption{Summary of the parameters calculated for Bondville, IL.}
    \label{tab:summary-bondville}
    \resizebox{\textwidth}{!}{%
    \begin{tabular}{lrrrrrrrrrrrrrrr}
        \toprule
        & \textbf{dt} & \textbf{zen} & \textbf{dw\_solar} & \textbf{diffuse} & \textbf{dw\_ir} & \textbf{dw\_dometemp} & \textbf{uvb} & \textbf{par} & \textbf{netsolar} & \textbf{totalnet} & \textbf{temp} & \textbf{rh} & \textbf{windspd} & \textbf{winddir} & \textbf{pressure} \\
        \midrule
        count & 156497 & 156497 & 156497 & 156497 & 156497 & 156497 & 156497 & 156497 & 156497 & 156497 & 156497 & 156497 & 156497 & 156497 & 156497 \\
        mean  & 14.62  & 56.97  & 356.75 & 152.49 & 347.08 & 290.72 & 51.01 & 155.32 & 283.19 & 226.03 & 15.86 & 75.09 & 4.87 & 180.19 & 991.33 \\
        std   & 1.48   & 15.94  & 271.79 & 111.52 & 52.54  & 9.17   & 51.85 & 114.18 & 216.05 & 188.51 & 8.71  & 16.48 & 2.78 & 95.69  & 5.76 \\
        min   & 10.98  & 20.41  & 0.10   & 1.10   & 209.50 & 272.14 & 0.10  & 0.70   & 0.30   & 0.10   & 0.10  & 15.10 & 0.10 & 0.10   & 969.40 \\
        25\%  & 13.48  & 44.89  & 119.10 & 71.50  & 314.90 & 282.33 & 10.50 & 55.70  & 96.10  & 63.20  & 7.90  & 64.40 & 2.80 & 104.00 & 988.20 \\
        50\%  & 14.73  & 58.75  & 293.60 & 118.10 & 355.00 & 292.74 & 30.80 & 130.50 & 232.10 & 175.20 & 18.30 & 77.80 & 4.30 & 189.50 & 991.30 \\
        75\%  & 15.88  & 69.88  & 562.40 & 206.50 & 389.40 & 298.26 & 77.30 & 243.60 & 442.50 & 357.70 & 23.00 & 88.40 & 6.50 & 250.90 & 994.90 \\
        max   & 16.98  & 85.00  & 1356.80 & 763.70 & 457.10 & 309.51 & 291.60 & 568.90 & 1112.40 & 1049.00 & 32.70 & 101.30 & 20.30 & 360.00 & 1009.00 \\
        \bottomrule
    \end{tabular}}
\end{table}

\subsection{Workflow Overview}
\Cref{fig:flowchart} depicts the complete workflow adopted for the development and evaluation of the solar-irradiation forecasting framework. The process begins with the acquisition of minute-resolution meteorological measurements from three geographically diverse SURFRAD stations (Bondville, Illinois; Penn State, Pennsylvania; and Desert Rock, Nevada). The raw observations are subjected to a systematic preprocessing stage, wherein physically implausible records (for example, placeholders such as --9999.9) are identified and removed. The remaining data are subsequently normalized to ensure statistical consistency across variables.

Following data cleansing, a feature selection procedure based on Random-Forest variable importance is employed to identify the most influential predictors of Global Horizontal Irradiance (GHI). The curated dataset is then partitioned into training (2018--2019) and validation (2020) subsets to facilitate unbiased model evaluation.

Three machine-learning approaches are subsequently implemented:
(i) Linear Regression (LR) as a baseline statistical model,
(ii) Extreme Gradient Boosting (XGB) as an ensemble tree-based learner, and
(iii) XGB with Genetic Algorithm (GA)--driven hyperparameter optimization, wherein GA operations of selection, crossover, and mutation iteratively refine the parameter set to achieve near-optimal model performance.

The final stage of the workflow involves quantitative performance assessment of all models on the independent validation set using established metrics---mean squared error (MSE), mean absolute error (MAE), explained variance, and predictive accuracy. The diagram therefore encapsulates the complete end-to-end pipeline, from data acquisition and preprocessing through feature engineering, model training and GA-based optimization, culminating in rigorous validation and comparative performance analysis.

\section{Results and Discussion}
The study employed three different machine learning models LR,XGB, and a GA based approach to predict solar irradiation using datasets from 2018 and 2019 for training and testing purposes, respectively. The XGB model was implemented using four different parameter sets to identify the optimal configuration for solar irradiation prediction, while the GA model was explored under three configurations with varying numbers of generations such as: 10, 20 and 50. 

The overview of the entire work flow is depicted in \Cref{fig:flowchart}, which shows input data sets, data preprocessing, feature selection, training, testing and validation of data. The flow shows on the evaluation and prediction of the model. The XGB is further processed with GA using different generations to improve the model further as can be seen in the flow chart, thereby producing an enhanced model predicting with higher accuracy. 

A cross-site comparison of model performance is provided in \Cref{tab:accuracy}. For each station (Bondville, Desert Rock, Penn State), we report MAE, explained variance, and overall accuracy for the LR baseline, XGB, and GA-optimized XGB under both Train--Test and Validation splits.

Among these, the LR model exhibited the lowest performance, with an accuracy of about $95.5\%$ and a mean absolute error (MAE) of $14.73$. Although the XGB model improved upon LR by achieving an accuracy of roughly $98.5\%$, it still did not match the performance of the GA-enhanced model. The GA model configured with 10 generations produced the best results on the test dataset, attaining an accuracy of $99\%$ with a significantly lower MAE of $2.74$. On the validation set, the GA approach also showed strong performance, achieving an accuracy of approximately $97.75\%$ and a MAE of $7.45$.

\begin{table}[!htbp]
    \centering
    \caption{Summary of the accuracy, MAE, and variance for the three stations.}
    \label{tab:accuracy}
    \resizebox{\textwidth}{!}{%
    \begin{tabular}{lccc ccc ccc}
        \toprule
        \multirow{2}{*}{\textbf{Model Comparison}} &
        \multicolumn{3}{c}{\textbf{Bondville-IL}} &
        \multicolumn{3}{c}{\textbf{DesertRock-NV}} &
        \multicolumn{3}{c}{\textbf{PennState-PA}} \\
        \cmidrule(lr){2-4}\cmidrule(lr){5-7}\cmidrule(lr){8-10}
        & \textbf{LR} & \textbf{XGB-100} & \textbf{GA 10}
        & \textbf{LR} & \textbf{XGB-100} & \textbf{GA 10}
        & \textbf{LR} & \textbf{XGB-100} & \textbf{GA 10} \\
        \midrule
        \multicolumn{10}{l}{\emph{Train--Test}} \\
        MAE      & 14.73 & 5.39 & 4.64 & 12.09 & 5.30 & 4.58 & 3.89 & 3.69 & 3.08 \\
        Variance & 88.19 & 97.93 & 98.42 & 90.13 & 98.00 & 98.47 & 98.94 & 99.01 & 99.28 \\
        Accuracy (\%) & 95.55 & 98.41 & 98.64 & 96.16 & 98.32 & 98.55 & 98.81 & 98.61 & 99.09 \\
        \midrule
        \multicolumn{10}{l}{\emph{Validation}} \\
        MAE      & 14.18 & 7.69 & 7.45 & 13.03 & 12.68 & 12.92 & 6.07 & 5.51 & 5.42 \\
        Variance & 88.23 & 95.70 & 95.95 & 89.85 & 89.43 & 88.81 & 97.01 & 97.91 & 97.96 \\
        Accuracy (\%) & 95.63 & 97.67 & 97.74 & 95.95 & 96.04 & 95.96 & 98.15 & 98.30 & 98.33 \\
        \bottomrule
    \end{tabular}}
\end{table}

Performance comparisons across different meteorological stations Bondville, IL; Desert Rock, NV; and Penn State, PA show that the GA model consistently outperformed both XGB and LR in the validation phase. For instance, while XGB generally provided better results than LR, an exception was observed at Penn State, PA, where LR marginally outperformed XGB by $0.2\%$. Nevertheless, across all stations, GA not only achieved the highest accuracy but also demonstrated the lowest MAE; most notably, Penn State recorded an accuracy of $98.33\%$ with a MAE of $5.42$.

\begin{figure}[!htbp]
    \centering
    \includegraphics[width=0.8\linewidth]{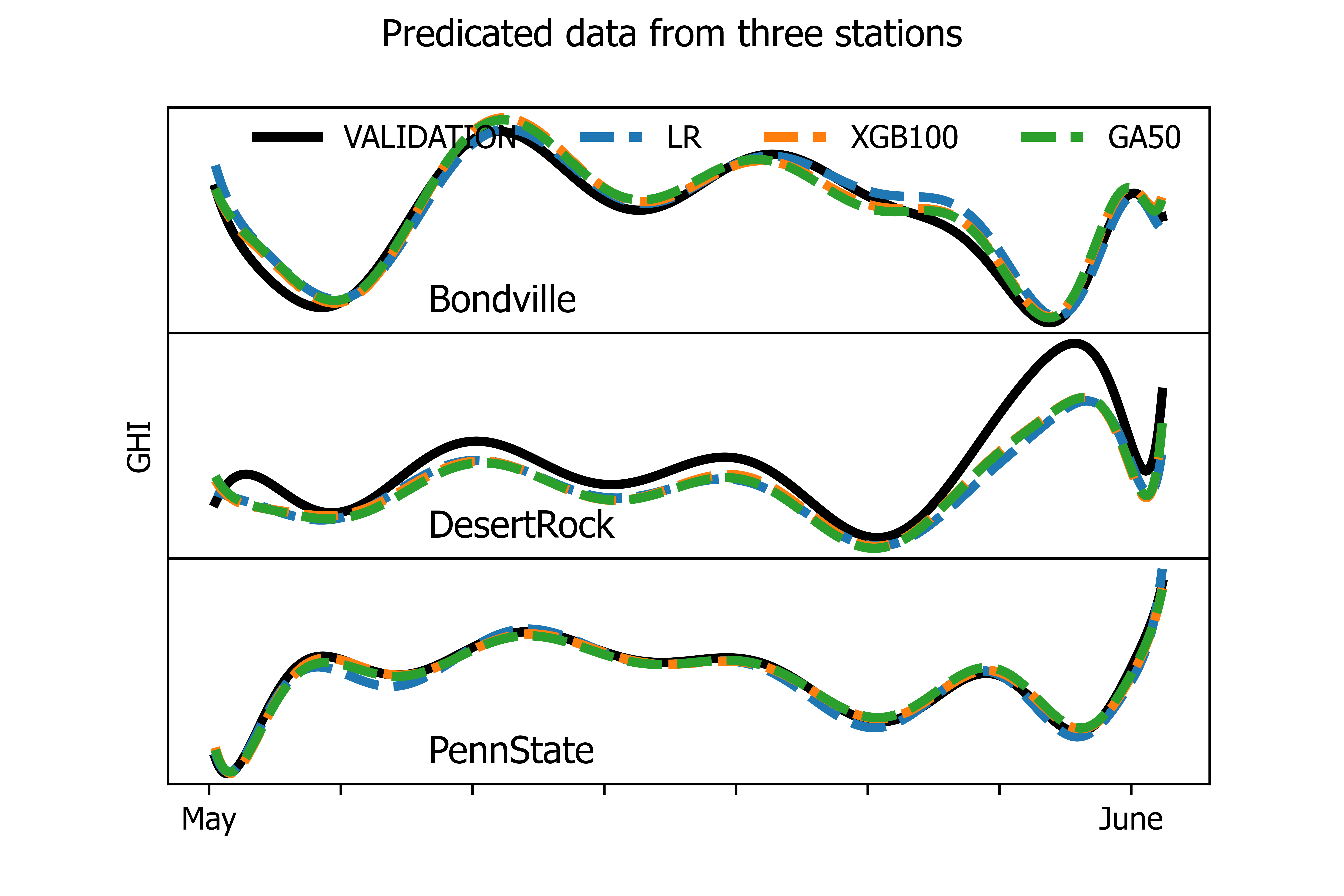}
    \includegraphics[width=0.8\linewidth]{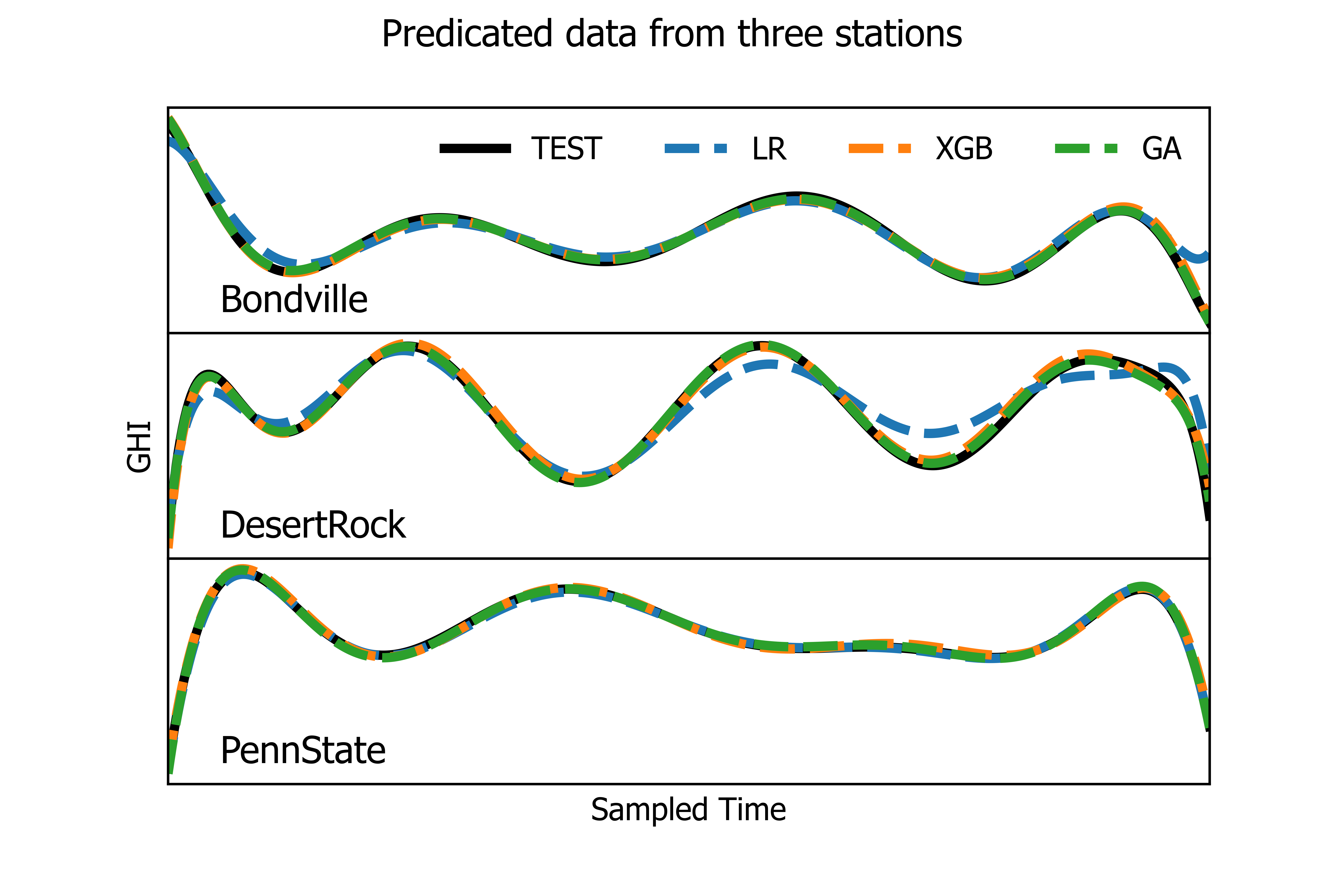}
    \caption{Test (top) and validation (bottom) plots for the three stations depicting the datasets compared with the predicted values from LR, XGB, and GA models for Bondville, IL; Desert Rock, NV; and Penn State, PA during May.}
    \label{fig:results}
\end{figure}

Beyond its high predictive accuracy, the GA model’s ability to automatically optimize hyperparameters confers significant practical advantages. This automation dramatically reduces the need for extensive manual tuning, saving both time and resources while minimizing human-induced errors in parameter selection. By enabling the model to quickly adapt to different datasets and changing environmental conditions, the GA approach is particularly well-suited for operational forecasting systems that require rapid updates and high reliability.

While the GA-optimized XGB model showed strong performance, there is still room for improvement by expanding the range of hyperparameters used during optimization. For example, testing more values for learning rate, tree depth, and regularization settings could help the model perform better across different weather conditions. Improving the way the genetic algorithm selects the best model such as, by considering both accuracy and consistency could also make the model more reliable.

Future work can also explore combining several GA-optimized models like XGB, LightGBM, and CatBoost into one ensemble model using methods like stacking or averaging. This can help reduce errors and improve performance in locations with very different climates. Adding more input features, such as past values of solar data, cloud cover, or satellite-based measurements, could also help the model better understand changes in sunlight and make more accurate predictions.

\section{Conclusion}
Data collected from three meteorological stations with diverse climatic conditions were used to evaluate the effectiveness of genetic algorithms in enhancing the accuracy of global solar irradiation forecasting. Machine learning techniques such as Linear Regression, Extreme Gradient Boosting, and Genetic Algorithm Optimization were applied and their prediction results compared. The findings demonstrate that the GA-optimized model outperforms the other techniques, delivering superior accuracy across all tested stations.

This study provides a basis for assessing the performance of different ML methods for solar irradiation prediction, despite the relatively small sample size. The consistent superiority of the GA approach suggests that automated hyperparameter optimization can significantly improve model performance, making it a promising tool for operational forecasting. Future research should aim to expand the sample size by incorporating data from additional meteorological centers with varying climatic conditions. Furthermore, increasing the number of input parameters and refining the existing parameter set could further enhance the predictive capability of the model.


\end{document}